\pdfoutput=1

\documentclass[11pt]{article}

\usepackage[]{acl}
\usepackage{times}
\usepackage{url}
\usepackage{latexsym}
\usepackage{amsmath,amssymb,amsfonts}
\usepackage{algorithmic}
\usepackage{graphicx}
\usepackage{textcomp}
\usepackage{kotex}
\usepackage{booktabs}
\usepackage{multirow}
\usepackage{arydshln}
\usepackage{tabularx}
\usepackage{color}
\usepackage{amssymb}
\usepackage{mathtools}
\usepackage{makecell}
\usepackage{hyperref}
\usepackage{appendix}

\usepackage[T1]{fontenc}

\usepackage[utf8]{inputenc}

\usepackage{microtype}

\title{
$\mathbf{GRASP}$: Guiding model with RelAtional Semantics using Prompt\\
for Dialogue Relation Extraction
}

\author{Junyoung Son\thanks{~~These authors have equally contributed to this work}, Jinsung Kim\footnotemark[1], Jungwoo Lim\footnotemark[1], Heuiseok Lim\thanks{~~Corresponding author} \\
Computer Science and Engineering, Korea University \\
Republic of Korea \\
\texttt{\{s0ny,jin62304,wjddn803,limhseok\}@korea.ac.kr}
}

\begin{document}
\maketitle
\begin{abstract}
The dialogue-based relation extraction (DialogRE) task aims to predict the relations between argument pairs that appear in dialogue. Most previous studies utilize fine-tuning pre-trained language models (PLMs) only with extensive features to supplement the low information density of the dialogue by multiple speakers. To effectively exploit inherent knowledge of PLMs without extra layers and consider scattered semantic cues on the relation between the arguments, we propose a \textbf{G}uiding model with \textbf{R}el\textbf{A}tional \textbf{S}emantics using \textbf{P}rompt ($\mathbf{GRASP}$). We adopt a prompt-based fine-tuning approach and capture relational semantic clues of a given dialogue with 1) an argument-aware prompt marker strategy and 2) the relational clue detection task. In the experiments, $\mathbf{GRASP}$ achieves state-of-the-art performance in terms of both F1 and F1$_c$ scores on a DialogRE dataset even though our method only leverages PLMs without adding any extra layers.
\end{abstract}

\section{Introduction}
\label{intro}
\begin{figure*}[ht]
    \centering
    \includegraphics[scale=0.22]{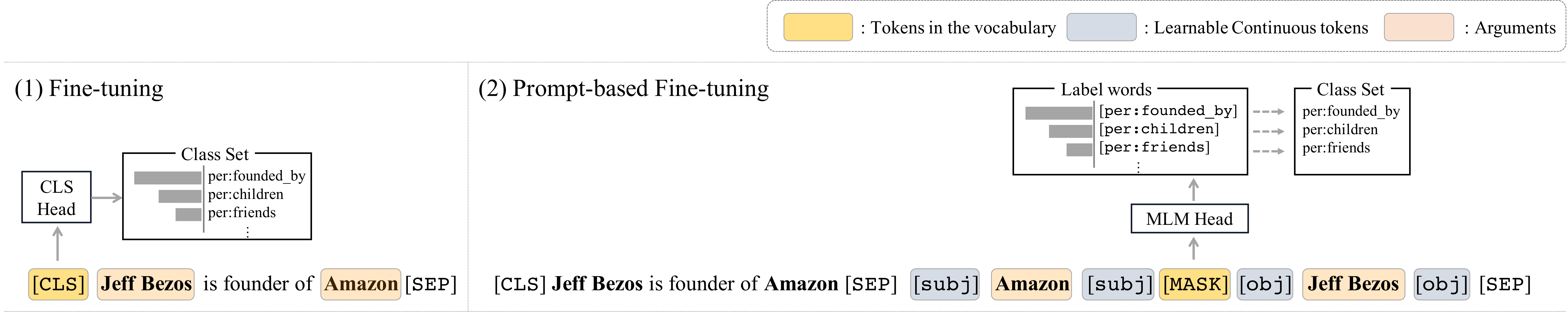}
    \caption{An illustration of (1) standard fine-tuning approach, and (2) prompt-based fine-tuning approach. In prompt-based fine-tuning approach, a set of \textit{label words} are mapped into a \textit{class set} by a certain mapping function.\label{fig:finetune_prompt}}
\end{figure*}

The relation extraction (RE) task aims to extract semantic relations from unstructured text such as a sentence, a document, or even a dialogue. RE plays a critical role in information extraction and knowledge base construction  as it can extract structured relational information \citep{ji2010overview,swampillai2010inter}. However, the utilization of sentence-level RE in a conversational setting is limited because numerous relational facts appear across multiple sentences with more than one speaker in a dialogue~\citep{yao2021codred}. Thus, the dialogue-based relation extraction (DialogRE) task, which includes argument pairs and their corresponding relations, has been proposed to encourage building a model that captures the underlying semantic spread in the dialogue, as presented in Table \ref{tab:dialogRE} \citep{yu-etal-2020-dialogue}. 
\begin{table}[t]
\centering
\scalebox{0.69}{
\begin{tabular}{@{}llll@{}}
\toprule
\multicolumn{4}{l}{\textbf{Dialogue}}                                                                      \\ \midrule
\multicolumn{4}{l}{\begin{tabular}[c]{p{10cm}}
1~~S1: Hey guys! Hey!\\
2~~S2: Hey \textbf{Pheebs}, guess who we saw today.\\
3~~\textbf{S3}: Ooh, ooh, fun! Okay... um, Liam Neeson.\\
...\\
7~~\textbf{S3}: The woman who cuts my hair!\\
8~~S4: Okay, look, this could be a really long game.\\
9~~S5: Your \underline{sister} \textbf{Ursula}.\\
10~~\textbf{S3}: Oh, really.\\
11~~S5: Yeah, yeah, she \underline{works over at} that place, uh...\\
12~~\textbf{S3}: \textbf{Rift’s}. Yeah, I know.\\
...\\
17~~S6: Um, \textbf{Pheebs}, so, you guys just don’t get along?\\
18~~\textbf{S3}: It’s mostly just dumb \underline{sister} stuff, you know, I mean, like, everyone always thought of her as the pretty one, you know... ...\\ 
\end{tabular}}               \\ \midrule
                             & \multicolumn{1}{r}{\textbf{Argument pair}} & \textbf{Trigger}      & \textbf{Relation Type}             \\
\multirow{2}{*}{\textbf{R1}} & (Pheebs, PER)           & \multirow{2}{*}{sister} & \multirow{2}{*}{per:siblings}      \\
                             & (Ursula, PER)                       &                       &                                    \\ \cdashline{1-4}[0.8pt/1.2pt]
\multirow{2}{*}{\textbf{R2}} & (S3, PER)           & \multirow{2}{*}{none} & \multirow{2}{*}{per:alternate\_names}      \\
                             & (Pheebs, PER)                       &                       &                                    \\ \cdashline{1-4}[0.8pt/1.2pt]
\multirow{2}{*}{\textbf{R3}} & (Rift's, ORG)           &  \multirow{2}{*}{works over at} & \multirow{2}{*}{per:employees\_or\_members} \\
                             & (Ursula, PER)                       &                       &                                    \\ \bottomrule
\end{tabular}
}\caption{Example of DialogRE data. The arguments are bold, and the triggers are underlined. The arguments and triggers are scattered throughout the dialogue, which leads to low information density. The triggers determine the direction of the proper relation indirectly. \label{tab:dialogRE}}
\end{table}

In previous studies where state-of-the-art (SoTA) performance is achieved on DialogRE benchmarks, fine-tuning is employed on pre-trained language models (PLMs) \citep{lee-choi-2021-graph,ijcai2021-535}, such as BERT \citep{devlin-etal-2019-bert} and RoBERTa \citep{liu2019roberta}. As fine-tuning requires the addition of extra layers on top of the PLMs and the training objectives are different from those used in the pre-training phase, PLMs cannot effectively exploit their learned knowledge in the downstream task, resulting in less generalized capability \citep{chen2021knowprompt}.

To effectively utilize knowledge from PLMs, several studies involving prompt-based fine-tuning have been conducted. They employ the PLM directly as a predictor and completing a cloze task to bridge the gap between pre-training and fine-tuning~\citep{gao2020making,han2021ptr}.
As presented in Figure \ref{fig:finetune_prompt}, prompt-based fine-tuning treats the downstream task as a masked language modeling (MLM) problem by directly generating the textual response to a given template. In specific, prompt-based fine-tuning updates the original input on the basis of the template and predicts the \textit{label words} with the \texttt{[MASK]} token. Afterwards, the model maps predicted \textit{label words} to corresponding task-specific class sets.

However, the prompt-based fine-tuning approach is still not sufficient in terms of performance compared with the fine-tuning-based approach~\citep{han2021ptr,chen2021knowprompt}. We attribute this phenomenon to the following properties of conversation: higher person-pronoun frequency~\citep{wang2011pilot} and lower information density~\citep{biber1991variation} by multiple speakers\footnote{In the DialogRE dataset, 65.9\% of relational triples involve arguments that never appear in the same utterance, demonstrating that multi-turn reasoning plays an important role.}. Therefore, a prompt-based fine-tuning approach that collects the sparse semantics in the dialogue is required to understand relation between the arguments.

We propose a method named \textbf{G}uiding model with \textbf{R}el\textbf{A}tional \textbf{S}emantics using \textbf{P}rompt ($\mathbf{GRASP}$) for DialogRE. To maximize the advantages of the prompt-based fine-tuning approach for the DialogRE task, we suggest an argument-aware prompt marking (APM) strategy and a relational clue detection (RCD) task. The APM strategy guides the model to the significant arguments scattered in the dialogue by carefully considering arguments. For our APM strategy, we conduct empirical study based on the diverse marker types to validate our APM strategy. Along with the APM strategy, the suggested RCD task with a training objective leads the model to pay attention to significant relational clues. Specifically, the model is trained to determine whether each token in a dialogue belongs to a subject, object, or trigger. As a result, the PLM is trained on RCD and MLM jointly. In the experiments, our method achieves SoTA performance at a significant level in the DialogRE task for both the full-shot and few-shot settings. Only PLMs are employed without the addition of an extra layer as a predictor, and $\mathbf{GRASP}$ exhibits a higher performance than other baselines. The significant performance improvement indicates that attending to significant semantic clues guides the PLMs to predict the correct relation with its inherent knowledge in both full-shot and few-shot settings. Moreover, we provide ablation studies and qualitative analysis on the robustness of $\mathbf{GRASP}$. 


Our contributions are as follows:
\begin{itemize}
\item {We adopt a prompt-based fine-tuning approach to utilize a PLM's inherent knowledge directly for dialogues with relatively low information density.}

\item {We introduce an APM strategy and a RCD task that guide PLMs on the significant relational clues, which are semantic information to predict relations.}

\item {We demonstrate that our proposed method achieves SoTA performance on the DialogRE task in both full-shot and few-shot settings. }

\item {We conduct ablation studies and qualitative analysis to validate the robustness of $\mathbf{GRASP}$.}

\end{itemize} 
The remainder of this paper is organized as follows. In Section \ref{method}, we present the entire process of our method in detail. The experimental setup and the results are explained in Section \ref{sec:experiments}. The further analyses is provided in Section \ref{analysis_section}, and Section \ref{conclusion} presents the conclusions. Appendix \ref{appendix:related_works} provides related works including the prompt-based learning, and the DialogRE. 

\section{Related Works\label{appendix:related_works}}
\paragraph{Prompt-based learning}
Prompt-based learning is a method of reducing the gap between the pre-training objective and that of fine-tuning. For example, language models such as BERT ~\citep{devlin-etal-2019-bert} use masked language modeling (MLM) objective in pre-training phase where the model fills the \texttt{[MASK]} token whereas the model trains without \texttt{[MASK]} token in fine-tuning phase by adding an extra classifier layer.
As a result, the discrepancy between training objectives prevents PLM from leveraging knowledge acquired from the pre-training enormous corpus~\citep{chen2021knowprompt}. 
Also, prompt-based learning shows better performance than fine-tuning especially in the few-shot setting~\cite{gao2020making,schick-schutze-2021-exploiting,li2021prefix,liu2021gpt}. 


\paragraph{DialogRE}
Recent studies on the DialogRE dataset show a tendency to fine-tune the PLM with task-specific objectives and use the model with high-complexity \citep{xue2021gdpnet,ijcai2021-535,lee-choi-2021-graph}. In detail, \citet{lee-choi-2021-graph} shows considerable performance with the contextualized turn representations from the diverse type of nodes and edges. Moreover, task-specific objectives of fine-tuning lead to a gap between pre-training and fine-tuning.

To overcome the limitation, \citet{han2021ptr} utilizes multiple \texttt{[MASK]} tokens for each argument and the relation with logical rules by concentrating subject and object in DialogRE. There also exists an approach that incorporates potential knowledge contained in relation labels into prompt construction with trainable virtual type words and answers words. This approach also carefully initializes the virtual tokens with implicit semantic words and employs prior distributions estimated from the data~\citep{chen2021knowprompt}.





\paragraph{}Despite the prompt-based approach's high potential, few prompting studies sufficiently consider low information density and difficulty of capturing intrinsic relational information of the data between the argument pair of dialogue relation extraction task. We focus on building a light model with prompt-based fine-tuning with implicit semantic information of the relation which alleviates the sparsity problem.




\begin{figure*}[ht]
    \centering
    \includegraphics[scale=0.38]{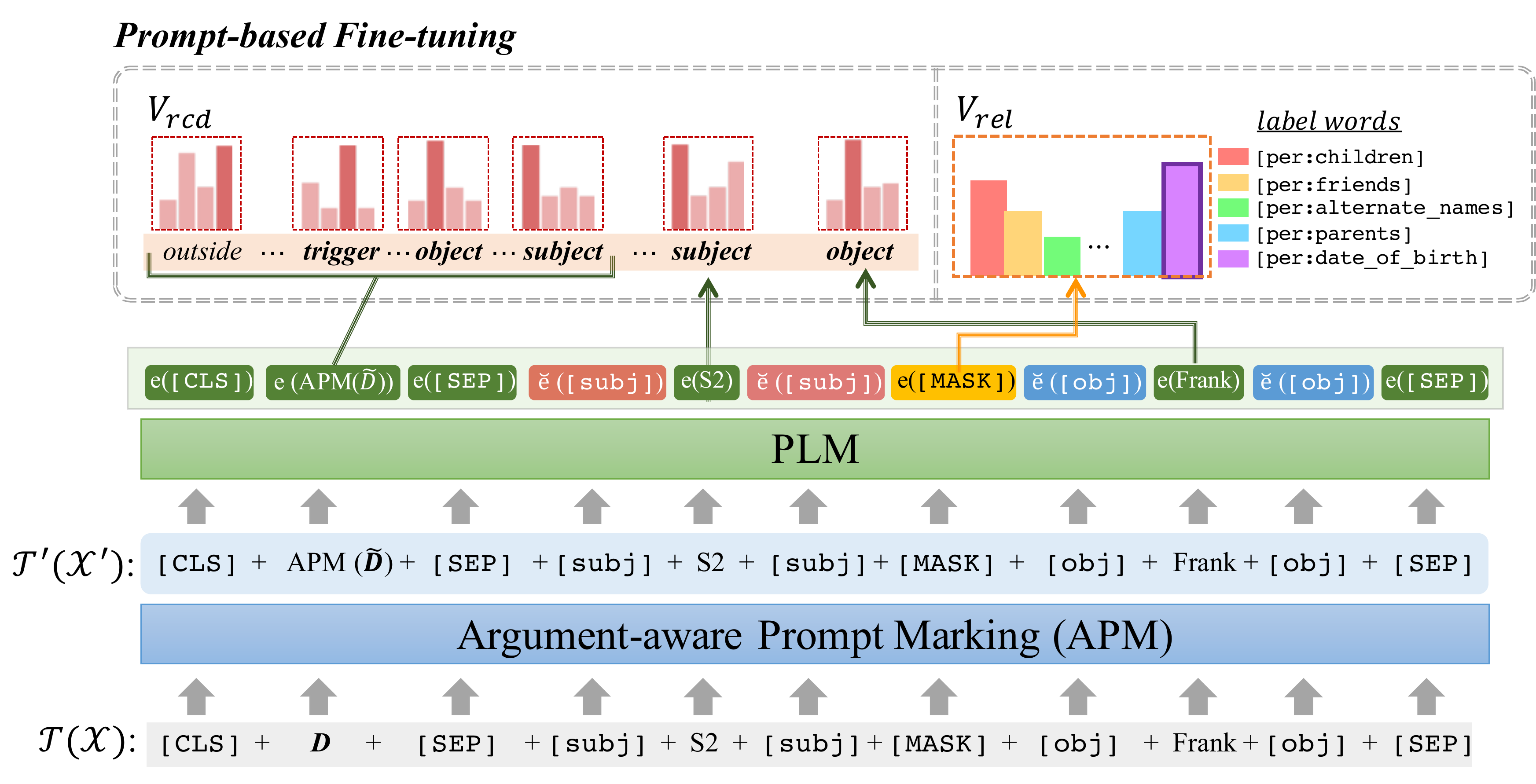}
    \caption{The overall model architecture of $\mathbf{GRASP}$. By formalizing specific tasks as MLM tasks, the model predicts answers with the tokens from the model's vocabulary; for example, the \textit{label words} of $V_{rel}$ and $V_{rcd}$ are from the model's vocabulary.  \label{fig:model_overview}}
\end{figure*}

\section{Methodology\label{method}}
An overview of $\mathbf{GRASP}$ is illustrated in Figure \ref{fig:model_overview}. First, an input with a prompt template is constructed by using the APM strategy. Then, the PLM receives the constructed input for prompt-based fine-tuning and estimates probability distributions of the model's vocabulary by using contextualized representations of the PLM. The RCD task lets the model predict the relational clue type of each token, and the model takes the \texttt{[MASK]} representation to predict a final relation in the MLM task. In other words, our model is trained through multitask learning to encourage mutual communication between relational clues and the final relation for the argument pair.

\subsection{Problem Formulation}
\label{subsec:problem_formulation}
Each example $\mathcal{X}$ includes dialogue $D$, subject $a_1$, and object $a_2$. Note that $D$ denotes $\{s_1:u_1, s_2:u_2,\dots,s_N:u_N\}$, where $s_n$ is the $n$-th speaker and $u_n$ is the corresponding utterance. Given $\mathcal{X}=\{D,a_1,a_2\}$, the goal of DialogRE is to predict relation $y$ between arguments $a_1$ and $a_2$ by leveraging $D$. To describe the DialogRE task in terms of prompt-based fine-tuning, a template function, $\mathcal{T}(\cdot)$, is defined to map each example to $\mathcal{X}_{prompt}=\mathcal{T}(\mathcal{X})$. 
A \texttt{[MASK]} is inserted into $\mathcal{X}_{prompt}$, and used to predicting the label words of relation $y$. The formulation of $\mathcal{T}(\mathcal{X})$ is as follows:
\begin{equation}\label{eq:prompt_design}
\scalebox{0.88}{$
\begin{aligned}
     \mathcal{T}(\mathcal{X}) = 
    \text{``}&\texttt{[CLS]} D \texttt{[SEP]} \texttt{[subj]} a_1 \texttt{[subj]} \\
    &\texttt{[MASK]} \texttt{[obj]} a_2 \texttt{[obj]} \texttt{[SEP]}\text{''}.
\end{aligned}
$}
\end{equation}
Based on the structure of $\mathcal{T}(\cdot)$, we construct our template function, $\mathcal{T}'(\cdot)$, 
by applying two steps: transformation of $D$ to $D'$ with an argument-aware prompt marker, described in \ref{subsection:encoding}, and prompt initialization for \texttt{[subj]} and \texttt{[obj]}, explained in \ref{subsection:prompt}. Subsequently, we introduce the RCD task in \ref{subsection:rcd} to train the model by utilizing $\mathcal{T}'(\cdot)$ with a multitask training strategy on MLM, as detailed in \ref{subsec:model_training}.

\subsection{Argument-aware Prompt Marker}
\label{subsection:encoding}

We propose an APM strategy that considers both speaker and non-speaker arguments. 
In previous studies, a dialogue is encoded by focusing on speaker information~\citep{lee-choi-2021-graph,yu-etal-2020-dialogue,chen2021knowprompt} without focusing non-speaker arguments. However, in the DialogRE dataset, approximately 77.4\% of relation triples include at least one non-speaker argument, implying that consideration of non-speaker arguments is also inevitable to enhance the model's argument-awareness.

Our methods is inspired by the previous works of \citet{soares2019matching} and \citet{han2021ptr}; the former addresses the importance of entity markers regarding recognition of the entity position, and the latter improves the model performance by using specific additional punctuation in the model's original vocabulary as the entity marker. 
Accordingly, we insert the argument-aware prompt marker token, \texttt{[p]}, as an entity marker. The argument-aware prompt markers allow our model to obtain informative signs to determine which token is the indicative component for relation prediction.
We initialize the feature of \texttt{[p]} as the embedding of the space token in the vocabulary of the model. Our empirical experiments reveal that the space token can perceive the start position of the arguments. Consequently, our prompt marker enhances the model to discriminate which part of the dialogue plays a critical role in predicting a relation.

Using the proposed argument-aware prompt marker, we strengthen the token replacement method of BERT${_s}$~\citep{yu-etal-2020-dialogue}. 
Given example $\mathcal{X}$, BERT${_s}$ constructs $\tilde{\mathcal{X}} = \{\tilde{D},\tilde{a_1},\tilde{a_2}\}$, where $\tilde{D} = \{ \tilde{s}_1:u_1, \tilde{s}_2:u_2, \dots, \tilde{s}_N:u_N \}$ and $\tilde{s_n}$ is
\begin{equation}
{
    \tilde{s_n} = \begin{cases} 
    \texttt{[S$_1$]} & $if$\ s_n=a_1 \\
    \texttt{[S$_2$]} & $if$\ s_n=a_2 \\
    s_n & $otherwise.$
    \end{cases}
} 
\end{equation} \label{equation:berts}
\texttt{[S$_1$]} and \texttt{[S$_2$]} are special tokens for speakers. In addition, $\tilde{a}_m$ ($m\in\{1,2\}$) is defined as \texttt{[S$_m$]} if $\exists~n~(s_n=a_m)$ and $a_m$ otherwise. 

Even though BERT${_s}$ prevents the model from overfitting and demonstrates higher generalization capacity~\citep{yu-etal-2020-dialogue}, BERT${_s}$ does not consider non-speaker arguments. To explore the disregarded arguments, the APM strategy expands BERT$_s$ by considering both types of arguments in the DialogRE task.
We define function \texttt{APM($\cdot$)} that encodes an utterance by inserting \texttt{[p]} in front of each argument token. Given dialogue $\tilde{D}$ constructed by using BERT$_s$, we construct $D' = \{\tilde{s}_1:u'_1, \tilde{s}_2:u'_2, \dots, \tilde{s}_N:u'_N\}$ by applying \texttt{APM}$(u_n)$ for each utterance in $\tilde{D}$. 
Consequently, the APM strategy constructs $\mathcal{X}' = \{D',\tilde{a}_1, \tilde{a}_2\}$ based on $\mathcal{\tilde{X}}$.
\begin{equation}
{
    \scalebox{0.85}{$u'_n = \begin{cases}
    \texttt{APM}(u_n) & $if$~\exists~m~(\tilde{a}_m \in u_n)~(m \in \{1,2\}) \\
    u_n & $otherwise$
    \end{cases}$}
}
\end{equation}
For instance, $\texttt{APM}(\cdot)$ encodes the text ``I am Tom Gordon'' to \texttt{[I, am, [p], Tom, Gordon]}, inserting \texttt{[p]} in front of ``Tom Gordon.'' 

\subsection{Prompt Construction}
\label{subsection:prompt}
We update the constructed input $\mathcal{X}'$ with a template function and conduct a deliberate initialization. We utilize the prior distribution of argument types for initialization inspired by the study conducted by \citet{chen2021knowprompt}. Prompt tokens \texttt{[subj]} and \texttt{[obj]} are used to inject argument-type information. We define the argument-type set, $\mathcal{AT}=\{$``PER,'' ``ORG,'' ``GPE,'' ``VALUE,'' ``STRING''$\}$, using the types pre-defined in the dataset as depicted in Table \ref{tab:dialogRE}. We calculate the distributions of argument types $\phi^{\texttt{[subj]}}$ and $\phi^{\texttt{[obj]}}$ over $\mathcal{AT}$ by using frequency statistics.
We aggregate each argument type, $at\in\mathcal{AT}$, with the corresponding prior distribution to initialize prompt tokens \texttt{[subj]} and \texttt{[obj]}. The specific initialization equations are as follows:
\begin{equation}
    \scalebox{0.9}{$
    \begin{aligned}
        \check{e}(\text{\texttt{[subj]}}) & = \sum_{at~\in \mathcal{AT}}{ \phi^{\texttt{[subj]}}_{at} \cdot e(at) } \\
        \check{e}(\texttt{[obj]}) & = \sum_{at~\in \mathcal{AT}}{ \phi^{\texttt{[obj]}}_{at} \cdot e(at) },
    \end{aligned}
    $}
\end{equation}
where $e(\cdot)$ is the embedding from the PLM of an input token and $\check{e}(\cdot)$ is the initialized embedding of the prompt token.
Suppose the subject has prior distribution, $\phi^{\texttt{[subj]}}=\{$``PER''$:0.5, $’`ORG''$:$0.5, ``GPE'':$0.0, $``VALUE'':$0.0, $``STRING'':$0.0$\}.  The initial embedding of the $\texttt{[subj]}$ token can be calculated as a weighted average, i.e., $\check{e}(\texttt{[subj]}) = 0.5 \cdot e($``PER''$) + 0.5 \cdot  e$(``ORG''$)$.

Consequently, we can formalize $\mathcal{T}'(\cdot)$, which converts $\mathcal{X}'$ to $\mathcal{X}'_{prompt}$, where $\mathcal{X}_{prompt}'$ is an argument-enhanced input example, by using the APM strategy and prompt construction with deliberate initializations, i.e., $\mathcal{X}'_{prompt} = \mathcal{T}'(\mathcal{X}')$.
Then, the final input structure for prompt-based fine-tuning is as follows:
\begin{equation}
    \scalebox{0.88}{$
    \begin{aligned}
    \mathcal{T'}(\mathcal{X}')=\text{``}&\texttt{[CLS]}D'\texttt{[SEP]}\texttt{[subj]}\tilde{a}_1\texttt{[subj]} \\
    &\texttt{[MASK]}\texttt{[obj]}\tilde{a}_2\texttt{[obj]}\texttt{[SEP]}\text{''}.
    \end{aligned}
    $}
\end{equation}

In addition, for relation prediction by applying MLM, we also define $\mathcal{V}_{rel}$ as a set of \textit{label words} in the model's vocabulary as illustrated in Figure \ref{fig:model_overview}. In detail, we utilize its metadata for the initialization of each relation representation to inject its semantics. For instance, we add a special token to the model's vocabulary, \texttt{[per:date\_of\_birth]} as a \textit{label word}, and initialize this token by aggregating the embeddings of the words in the metadata, i.e., \{``person,'' ``date,'' ``of,'' ``birth''\} for a class ``per:date\_of\_birth.''

\begin{table*}[ht]
\centering
\scalebox{0.825} {
\begin{tabular}{lcccc}
\toprule
\multicolumn{5}{c}{\textit{Full-shot Setting}}                                                                                                                   \\ \midrule
\multicolumn{1}{l|}{\multirow{2}{*}{Method}} & \multicolumn{2}{c|}{V1}                             & \multicolumn{2}{c}{V2}                             \\
\cline{2-5}
\multicolumn{1}{l|}{}                        & \multicolumn{1}{c}{F1} & \multicolumn{1}{c|}{F1$_c$} & \multicolumn{1}{c}{F1} & \multicolumn{1}{c}{F1$_c$} \\ \midrule
\multicolumn{5}{c}{Fine-tuning based approach}                                                                                                          \\ \midrule
\multicolumn{1}{l|}{BERT$_s$ \citep{yu-etal-2020-dialogue}}                       & \multicolumn{1}{c}{61.2} & \multicolumn{1}{c|}{55.4} & \multicolumn{1}{c}{-} & \multicolumn{1}{c}{-}                 \\
\multicolumn{1}{l|}{RoBERTa$_s$ \citep{lee-choi-2021-graph}}                       & - & \multicolumn{1}{c|}{-}  & 71.3                        & 63.7 \\
\multicolumn{1}{l|}{Dual \citep{bai-etal-2021-semantic}}          & 67.3  & \multicolumn{1}{c|}{\underline{61.4}} & 67.1 & 61.1 \\
\multicolumn{1}{l|}{CoIn \citep{ijcai2021-535}}                       & \multicolumn{1}{c}{\underline{72.3}}  & \multicolumn{1}{c|}{-}  & -   &   -               \\
\multicolumn{1}{l|}{TUCORE-GCN \citep{lee-choi-2021-graph}}                       & \multicolumn{1}{c}{-} & \multicolumn{1}{c|}{-}  & \underline{73.1}  & \underline{65.9} \\ \midrule
\multicolumn{5}{c}{Prompt-based fine-tuning approach}                                                                                                   \\ \midrule
\multicolumn{1}{l|}{PTR \citep{han2021ptr}}                       & \multicolumn{1}{c}{63.2} & \multicolumn{1}{c|}{-}      & - & - \\
\multicolumn{1}{l|}{KnowPrompt \citep{chen2021knowprompt}}                       &   \multicolumn{1}{c}{68.6} & \multicolumn{1}{c|}{-}      & \multicolumn{1}{c}{-}  & \multicolumn{1}{c}{-}                       \\
\cdashline{1-5}[0.8pt/1.2pt]
\multicolumn{1}{l|}{$\mathbf{GRASP}_{base}$ (Our model)}                       & 69.2 & \multicolumn{1}{c|}{62.4}      &        69.0  & 61.7 \\
\multicolumn{1}{l|}{$\mathbf{GRASP}_{large}$ (Our model)}                       &  \textbf{75.1} (+2.8)  & \multicolumn{1}{c|}{\textbf{66.7} (+5.3)} &                       \textbf{75.5} (+2.4) & \textbf{67.8} (+1.9) \\ \bottomrule
\end{tabular}
} \caption{Performances of $\mathbf{GRASP}$ on test set of DialogRE. V1 and V2 represent the version of the dataset. The underlined scores are the previous SoTA performances. Subscript in parentheses represents advantages of $\mathbf{GRASP}$ over the best results of baselines (the underlined). Best results are bold.\label{tab:main_experiments}}
\end{table*}

\subsection{Relational Clue Detection task}\label{subsection:rcd}
To improve the understanding capability on relational clues by employing a prompt-based fine-tuning approach, we introduce a RCD task. We define a set of \textit{label words}, $\mathcal{V}_{rcd}=$\{\texttt{[subject]}, \texttt{[object]}, \texttt{[trigger]}, \texttt{[outside]}\}, and add to the model's vocabulary. 
Then, we construct a sequence of \textit{label words} for RCD, $\mathcal{C}_{rcd}$, by assigning each token in $\mathcal{X}'_{prompt}$ to the corresponding clue type word from $\mathcal{V}_{rcd}$.
For instance, $\mathcal{C}_{rcd}$ is constructed as follows: \{\texttt{[subject]}, \texttt{[trigger]}, \texttt{[object]}\} when the token sequence is given by $\{\textit{``Pheebs''}, \textit{``lives in''}, \textit{``LA''}\}$, where \textit{``Pheebs''} is a subject and \textit{``LA''} is an object argument, and \textit{``lives in''} is a trigger. 

The RCD task exploits the MLM head that is used to predict the \texttt{[MASK]} token.
Except for the \texttt{[MASK]} token, each token is sequentially tagged with $\mathcal{V}_{rcd}$ using the meta-data provided in the dataset.
In other words, the RCD task allows the model to identify which non-\texttt{[MASK]} tokens correspond to certain relational clue types. RCD supports the model in collecting scattered information from the entire dialogue by indicating where to focus in the dialogue to predict a relation. In this respect, the model pays considerably more attention to the semantic clues of the relation, such as triggers. Moreover, our model maintains a lightweight complexity by conducting the RCD task without an additional classifier.
The loss for the RCD task over each token $x \in \mathcal{X}'_{prompt}$ is aggregated as follows:
\begin{equation} \label{equ:rcd_loss}
    \scalebox{0.85}{$\begin{aligned}
        \mathcal{L}_{RCD} & = -\sum_{x \in \mathcal{X}'_{prompt}}{\log P(x=\mathcal{C}_{rcd}(x)|\mathcal{X}'_{prompt})}  \\
    \end{aligned}$}.
\end{equation}

\subsection{Model Training} \label{subsec:model_training}
Before the final relation-prediction, 
we mark the position of the trigger using the \texttt{[p]} token, i.e., the argument-aware prompt marker, based on the results of the RCD task.
Given example $\mathcal{X}'_{prompt}$, the model predicts the \textit{label words} of $\mathcal{V}_{rcd}$ for all non-\texttt{[MASK]} tokens. Subsequently, \texttt{[p]} is appended in front of the words predicted as a trigger to train the model to distinguish essential clues that encourage determining the relation. 

\paragraph{Multitask Learning}

$\mathcal{M}_{rel}:\mathcal{Y}\to\mathcal{V}_{rel}$ is a mapping function that converts a class set, $\mathcal{Y}$, into a set of \textit{label words}, $\mathcal{V}_{rel}$. 
For each input $\mathcal{X}'_{prompt}$, the purpose of MLM is to fill \texttt{[MASK]} with the relation \textit{label words} in the model's vocabulary.
As the model predicts the correct \textit{label word} at the position of \texttt{[MASK]}, we can formulate $p(y|x)=P(\texttt{[MASK]}=\mathcal{M}_{rel}(y)|\mathcal{X}'_{prompt})$.
The training objective of the relation prediction is to minimize
\begin{equation} \label{MLM_loss}
    \scalebox{0.87}{$
    \mathcal{L}_{REL} = -\log P(\texttt{[MASK]}=\mathcal{M}_{rel}(y)|\mathcal{X}'_{prompt}).
    $}
\end{equation}

To improve the model's ability to capture relational clues through the interaction between the MLM task for relation prediction and the RCD task,
we employ a multitask learning to train the $\mathbf{GRASP}$ model by using the joint loss expressed in Equations~(\ref{equ:rcd_loss}) and~(\ref{MLM_loss}). Therefore, the final learning objective is to minimize the joint loss, where $\lambda_{1}$ and $\lambda_{2}$ are hyperparameters.
\begin{equation} \label{final_loss}
    \scalebox{0.9}{$\mathcal{L}_{GRASP} = \lambda_{1} \cdot \mathcal{L}_{RCD} + \lambda_{2} \cdot \mathcal{L}_{REL}$}
\end{equation}

\section{Experiments\label{sec:experiments}}
\subsection{Experimental Setup}
For the base PLMs, RoBERTa-base and RoBERTa-large are adopted, as denoted by $\mathbf{GRASP}_{base}$ and $\mathbf{GRASP}_{large}$, respectively. The evaluation metrics are the F1 and F1$_c$ \citep{yu-etal-2020-dialogue} scores. F1$_c$ is an evaluation metric for supplementing the F1 score in a conversational setting and is computed by employing part of the dialogue necessary for predicting the relation between given arguments as input instead of the entire dialogue. The detailed settings for $\mathbf{GRASP}$ can be found in Appendix \ref{appendix:experiment_settings}.

In a full-shot setting, $\mathbf{GRASP}$ is compared with both fine-tuning-based and prompt-based fine-tuning approaches.
TUCORE-GCN~\citep{lee-choi-2021-graph} is a typical fine-tuning-based model using turn-level features with a graph convolution network~\citep{kipf2016semi}. CoIn~\citep{ijcai2021-535} employs utterance-aware and speaker-aware representations, 
and Dual~\citep{bai-etal-2021-semantic} models the relational semantics using abstract meaning representations~\citep{banarescu-etal-2013-abstract}. Moreover, PTR~\citep{han2021ptr} and KnowPrompt~\citep{chen2021knowprompt} are the prompt-based fine-tuning baseline models. In the few-shot setting, 8-, 16-, and 32-shot experiments were conducted based on LM-BFF~\citep{gao2020making} by using three different randomly sampled data. 

\subsection{Experimental Results}
\paragraph{Full-shot setting}

As presented in Table \ref{tab:main_experiments}, it is shown that $\mathbf{GRASP}_{large}$ surpasses all of the baseline models, including the current SoTA models, that is, CoIn and TUCORE-GCN. From the result in the full-shot setting, the baselines of the fine-tuning-based approach, such as TUCORE-GCN or CoIn, show better performance than those of the prompt-based fine-tuning approach. In particular, CoIn outperforms all of the other baselines, including PTR and KnowPrompt on V1, and TUCORE-GCN exhibits the best performance on V2. Interestingly, even though the performance of the prompt-based fine-tuning baselines is much lower than fine-tuning based models, our $\mathbf{GRASP}_{large}$ outperforms regardless of the way of training approach. 

In addition, $\mathbf{GRASP}_{large}$ shows its efficiency in conversational settings by exceeding all the baselines in terms of F1$_c$, thereby indicating that our method effectively overcomes the low information density of dialogues. These results imply that guiding the model to pay attention to relational clues with a prompt-based fine-tuning approach can be more effective than adding additional features and layers. 
The slightly low performance of $\mathbf{GRASP}_{base}$ is attributed to the gap in the model size; for example, TUCORE-GCN has 401M weight parameters with RoBERTa-large model, and $\mathbf{GRASP}_{base}$ has 125M weight parameters with RoBERTa-base.   Nevertheless, $\mathbf{GRASP}_{base}$ demonstrates 6.0\%p and 0.6\%p improvements compared with the other prompt-based fine-tuning baselines, i.e., PTR and KnowPrompt, respectively. 


\begin{table}[t]
\centering
\scalebox{0.766}{\begin{tabular}{lccc}
\toprule
\multicolumn{4}{c}{\textit{Few-shot Setting}}                                      \\ \midrule
\multicolumn{1}{l|}{\multirow{2}{*}{Method}} & \multicolumn{3}{c}{Shot}               \\ 
\cline{2-4}
\multicolumn{1}{c|}{}                        & \textit{K=8}         & \textit{K=16}        & \textit{K=32}        \\ \midrule
\multicolumn{4}{c}{Fine-tuning based approach} \\ \midrule
\multicolumn{1}{l|}{RoBERTa \citep{chen2021knowprompt}}             & 29.8        & 40.8        & 49.7        \\
\multicolumn{1}{l|}{TUCORE-GCN \citep{lee-choi-2021-graph}}             & 24.6        & 40.0        & 53.8        \\ \midrule
\multicolumn{4}{c}{Prompt-based fine-tuning approach} \\ \midrule
\multicolumn{1}{l|}{PTR \citep{han2021ptr}}                     & 35.5        & 43.5        & 49.5        \\
\multicolumn{1}{l|}{KnowPrompt \citep{chen2021knowprompt}}              & \underline{43.8}        & 50.8        & 55.3        \\
\cdashline{1-4}[0.8pt/1.2pt]
\multicolumn{1}{l|}{\textbf{$\mathbf{GRASP}_{base}$} (Our model)}             & \textbf{45.4} & \underline{52.0} & \underline{56.0} \\
\multicolumn{1}{l|}{\textbf{$\mathbf{GRASP}_{large}$} (Our model)}             & 36.0 & \textbf{55.3} & \textbf{62.6} \\
\bottomrule
\end{tabular} 
}\caption{Low-resource RE performance of F1 scores (\%) on different test sets. We use K = 8, 16, 32 (\# of examples per class) for few-shot experiments. Best
results are bold and the second place results are underlined.\label{table:few_shot}}
\end{table}

\paragraph{Few-shot setting}
As presented in Table \ref{table:few_shot}, $\mathbf{GRASP}$ still exhibits robust performance in few-shot settings. $\mathbf{GRASP}_{base}$ outperforms the baselines of both the fine-tuning and prompt-based fine-tuning methods regardless of the number of shots, demonstrating 20\%p or higher performance in the 8-shot setting compared with TUCORE-GCN and indicating that our method is more efficient than the fine-tuning method in a low-resource setting. $\mathbf{GRASP}_{base}$ also demonstrates improved performance compared with KnowPrompt in all-shot settings, indicating the effectiveness of the considerations on the properties of the dialogue with prompt-based fine-tuning. Except for the 8-shot setting, $\mathbf{GRASP}_{large}$ presents outstanding performance, achieving up to 15.3\%p of absolute improvement in the 16-shot setting. 
Although $\mathbf{GRASP}_{large}$ outperforms the fine-tuning-based models and PTR, the limited performance of $\mathbf{GRASP}_{large}$ in the 8-shot setting can be attributed to an insufficient number of examples.

We also observe that the fine-tuning-based models, such as TUCORE-GCN, perform at least 5.7\%p worse than the prompt-based fine-tuning models, such as PTR, in the 8-shot setting, indicating that the fine-tuning-based models may have difficulty in sufficiently capturing relational semantics when the data are extremely scarce. In particular, TUCORE-GCN indicates a 5.2\%p lower performance than the fine-tuned RoBERTa, indicating that the high complexity requires a larger amount of data than the other models.

\begin{table}[t]
\centering
\scalebox{0.775}{
\begin{tabular}{p{4.5cm}cccc}
\toprule
\multirow{2}{*}{Method}          & \multicolumn{2}{c}{Dev} & \multicolumn{2}{c}{Test} \\
& F1 & F1$_c$ & F1 & F1$_c$ \\ \midrule
$\mathbf{GRASP}_{base}$ & \textbf{70.3} & \textbf{63.3} & \textbf{69.0} & \textbf{61.7} \\ \cdashline{1-5}[0.8pt/1.2pt]
~~-\textit{APM}                 & 69.5        & 62.9       & 66.9        & 60.7        \\
~~-\textit{RCD} & 69.2 & 62.8 & 67.7 & 61.0 \\
~~-\textit{Prompt manual init.}& 68.1 & 61.9 & 65.9 & 59.8 \\
\bottomrule
\end{tabular}
}
\caption{
Ablation study on DialogRE dataset. \label{table:grasp_ablation}}
\end{table}
\begin{table*}[ht]
\centering
\scalebox{0.725}{
\begin{tabular}{l|p{15cm}|c}
\toprule
\textbf{Marker Type} & \textbf{Input Example} & \textbf{F1} \\
\midrule
Entity marker           & \texttt{{[}CLS{]}} \texttt{{[}E1{]}} \textbf{Frank} \texttt{{[}/E1{]}} lives in \texttt{{[}E2{]}} \textbf{Montauk} \texttt{{[}/E2{]}} . \texttt{{[}SEP{]}}                 & 65.9                           \\
Type marker           & \texttt{{[}CLS{]}} \texttt{[E1:PER]} \textbf{Frank} \texttt{[/E1:PER]} lives in \texttt{[E2:GPE]} \textbf{Montauk} \texttt{[/E2:GPE]} . \texttt{{[}SEP{]}}                 & 66.7                           \\
Punctuation marker (!)  & \texttt{{[}CLS{]}} ! \textbf{Frank} ! lives in ! \textbf{Montauk} ! . \texttt{{[}SEP{]}}                                               & 66.7                           \\
Punctuation marker (@) & \texttt{{[}CLS{]}} @ \textbf{Frank} @ lives in @ \textbf{Montauk} @ . \texttt{{[}SEP{]}}                                           & 65.3 \\
Punctuation marker (;) & \texttt{{[}CLS{]}} ; \textbf{Frank} ; lives in ; \textbf{Montauk} ; . \texttt{{[}SEP{]}}                                           & 67.1 \\
\cdashline{1-3}[0.8pt/1.2pt]
APM marker (front) & \texttt{{[}CLS{]}} {\texttt{[p]}} \textbf{Frank} lives in {\texttt{[p]}} \textbf{Montauk} . \texttt{{[}SEP{]}} & \textbf{69.0} \\
APM marker (surrounding) & \texttt{{[}CLS{]}} {\texttt{[p]}} \textbf{Frank} {\texttt{[p]}} lives in {\texttt{[p]}} \textbf{Montauk} {\texttt{[p]}} . \texttt{{[}SEP{]}} & 65.9 \\ 
\bottomrule
\end{tabular}
}
\caption{The performance based on the marker type. The arguments are bold. The embedding of \texttt{[p]} is initialized with the space token from the model's vocabulary. In type marker, \texttt{[E1:PER]} represents a start position of subject which has a person type and \texttt{[/E1:PER]} represents an end position of object that has the same type.\label{table:prompt_marker_analysis}}
\end{table*}

\paragraph{Ablation Study}
We conduct an ablation study to validate the effectiveness of the proposed modules. As shown in Table \ref{table:grasp_ablation}, each of the proposed modules improves the overall performance for both F1 and F1$_c$ settings. Without the APM strategy, the performance of $\mathbf{GRASP}_{base}$ decreases the F1 score by 0.8\%p and the F1$_c$ score by 0.4\%p on the development set, and the F1 score drops sharply by 2.1\%p and F1$_c$ by 1.0\%p on the test set. This result indicates that argument-awareness can be obtained through both speaker and non-speaker argument information. When the RCD task is excluded, the performance of $\mathbf{GRASP}_{base}$ decreases the F1 score by 1.1\%p and F1$_c$ by 0.5\%p on the development set and the F1 score by 1.3\%p and F1$_c$ by 0.7\%p for the test set. These results demonstrate that the RCD task alleviates the low information density of the dialogue by guiding the model to focus on relational clues.

In addition, the performance without the manual initialization of prompt construction of $\mathbf{GRASP}_{base}$ is reduced by 2.2\%p for the F1 score and 1.4\%p for the F1$_c$ score on the development set and by 3.1\%p for the F1 score and 1.9\%p for the F1$_c$ score on the test set. This result suggests that prompt construction is a basic step in training the model in the prompt-based fine-tuning case. The deliberate initialization of prompts is critical for modeling the tasks in an appropriate direction.

\begin{table*}[ht]
\centering
\scalebox{0.775}{
\begin{tabular}{ccccc}
    \toprule
    \multicolumn{5}{l}{\textbf{Dialogue}} \\ \midrule
    \multicolumn{5}{l}{\begin{tabular}[c]{p{19cm}}
S1: Hey!!\\
S2: Hey!\\
S1: Guess what. \textbf{Frank Jr.}, and \textbf{Alice} \underline{got married}!\\
S2: Oh my God!!\\
S1: And! And, they’re gonna have a baby! And! And, they want me to grow it for them in my uterus.\\
S3: My God!\\
S4: Are you serious?\\
S1: Yeah\\
S5: You’re really thinking about having sex with your brother?!\\
S1: Ewww! And "Oh no!" It’s—they just want me to be the surrogate. It’s her-it’s her egg and her sperm, and I’m-I’m just the oven, it’s totally their bun.\\
S5: Huh.
    \end{tabular}} \\ \midrule
    \multirow{2}{*}{\textbf{Argument pair}} & \multicolumn{1}{c|}{\multirow{2}{*}{\textbf{Ground Truth}}} & \textbf{RoBERTa}  & \multicolumn{2}{|c}{$\mathbf{GRASP}$} \\ & \multicolumn{1}{c|}{} & Predicted Relation & \multicolumn{1}{|c}{Predicted (\textit{subject, object, trigger})} & Predicted Relation \\ \midrule
    (Frank Jr., Alice) & \multicolumn{1}{c|}{\textit{per:spouse}} & unanswerable & \multicolumn{1}{|c}{(Frank Jr, Alice, \underline{got married})} & per:spouse \\
    (Alice, Frank Jr.) & \multicolumn{1}{c|}{\textit{per:spouse}} & per:siblings & \multicolumn{1}{|c}{(Alice, Frank Jr, \underline{got married})} & per:spouse \\
    \bottomrule
    \end{tabular}
}\caption{The qualitative analysis on the prediction of $\mathbf{GRASP}$ based on the comparison with the RoBERTa model. Predicted (\textit{subject, object, trigger}) is that $\mathbf{GRASP}$ predicted on RCD task.\label{table:rcd_qualitative_analysis}}
\end{table*}

\section{Analysis\label{analysis_section}}
\subsection{Analysis on marker type for APM}

To analyze argument-awareness regarding the types of markers, we conduct experiments on diverse prompt markers, as shown in Table \ref{table:prompt_marker_analysis}. The result reveals that considering argument types leads to performance improvement in the model. Punctuation marker ``;'' shows comparable performance among other punctuation markers, and ``!'' and ``@'' achieve a limited score. We presume that the higher frequency of ``;'' acts as a delimiter, which results in decent performance. 

We also observe that the APM marker (front) performs the best, with a 69.0\% F1 score among all other marker types. In addition, we conduct an experiment based on the position of the prompt marker, \texttt{[p]}, by comparing two versions of the APM marker: APM marker (front) and APM marker (surrounding). The APM marker (surrounding) display 3.1\%p lower performance than the APM marker (front). Based on these results, we empirically adopt the embedding initialization of our \texttt{[p]} prompt marker using the space token and located it in front of the arguments.

\subsection{Qualitative Analysis on $\mathbf{GRASP}$}
\label{subsec:qualitative}
Since we train $\mathbf{GRASP}$ attending on relational clues through the APM strategy and the RCD task, we further conduct analysis to investigate that the relational clues such as triggers contribute to predicting a relation between the arguments in a prompt-based manner, as shown in Table \ref{table:rcd_qualitative_analysis}. Specifically, we compare $\mathbf{GRASP}_{large}$ with a fine-tuned RoBERTa-large model to validate our method.

We observe that the fine-tuned RoBERTa model struggles to capture the symmetrical relations including the trigger. The fine-tuned RoBERTa model fail to capture relational clues, such as ``got married'', misleading the model into predicting an inappropriate relations for both symmetrical relations between ``Frank Jr.'' and ``Alice.'' ``got married'' is a critical cue to distinguish the relations between ``per:spouse'' and ``per:siblings'' because this phrase implies a romantic relationship in a dictionary definition. 
In contrast, $\mathbf{GRASP}$, which is trained using the RCD task and the APM strategy in a prompt-based manner, predicted the correct relations, capturing the correct relational clues including the trigger ``got married'' for both symmetrical argument pairs. This result presents the effectiveness of $\mathbf{GRASP}$ designed to guide the model on the relational clues, alleviating the difficulties of low information density in dialogues.
Additional examples demonstrating similar phenomena for the symmetrical relations are provided in the Appendix \ref{appendix:qualitative}.

\begin{table}[t]
\centering
\scalebox{0.775}{
\begin{tabular}{l|c|c}
\toprule
\multirow{1}{*}{Method}     & MELD & EmoryNLP  \\ \midrule
RoBERTa~\citep{liu2019roberta}   & \multicolumn{1}{c|}{62.0} & 37.3   \\
COSMIC~\citep{ghosal-etal-2020-cosmic} & \multicolumn{1}{c|}{65.2} & 38.1 \\
\makecell[l]{TUCORE-GCN\\\citep{lee-choi-2021-graph}} & \multicolumn{1}{c|}{65.4} & 39.2       \\
\cdashline{1-3}[0.8pt/1.2pt]
$\mathbf{GRASP}_{large}$ (Ours) & \multicolumn{1}{c|}{\textbf{65.6}} & \textbf{40.0} \\ \bottomrule
\end{tabular}
}\caption{Experimental results of $\mathbf{GRASP}_{base}$ on MELD and EmoryNLP tasks. \label{table:rcd_ablation}}
\end{table}

\subsection{Analysis on the applicability of $\mathbf{GRASP}$}
To demonstrate the robustness of our APM strategy and RCD task, we evaluated $\mathbf{GRASP}$ on MELD~\citep{poria-etal-2019-meld} and EmoryNLP~\citep{zahiri2018emotion} datasets, which are designed for emotion recognition in conversations (ERC). 
MELD~\citep{poria-etal-2019-meld} is a multimodal dataset collected from a TV show named Friends and consists of seven emotion labels and 2,458 dialogues with only textual modality. 
EmoryNLP~\citep{zahiri2018emotion} is also collected from Friends and comprises seven emotion labels and 897 dialogues. Each utterance in these datasets is annotated with one of the seven emotion labels. The weighted-F1 is calculated to evaluate the MELD and EmoryNLP datasets.

For baselines, we employ the fine-tuned RoBERTa~\citep{liu2019roberta}, COSMIC~\citep{ghosal-etal-2020-cosmic}, and TUCORE-GCN models~\citep{lee-choi-2021-graph}. 
COSMIC~\citep{ghosal-etal-2020-cosmic} uses RoBERTa-large as the encoder. It is a framework that models various aspects of commonsense knowledge by considering mental states, events, actions, and cause-effect relations for emotional recognition in conversations. 

As presented in Table \ref{table:rcd_ablation}, $\mathbf{GRASP}$ is applied to other dialogue-based tasks by alleviating the low information density of the given dialogue. In particular, the performance of $\mathbf{GRASP}$ surpasses that of TUCORE-GCN, which is the current SoTA model in DialogRE, and those of the baseline specialized on ERC tasks, such as COSMIC in both MELD and EmoryNLP.

\section{Conclusion\label{conclusion}}

In this paper, we proposed $\mathbf{GRASP}$, which is a method for guiding PLMs to relational semantics using prompt-based fine-tuning for the DialogRE task. We focused on alleviating the critical challenge in dialogues, that is, low information density, by effectively capturing relational clues. In $\mathbf{GRASP}$, we constructed prompts with deliberate initialization and suggested 1) an APM strategy considering both speaker and non-speaker arguments and 2) the RCD task, which guides the model to determine which token belongs to the relational clues. Experimental results on the DialogRE dataset revealed that $\mathbf{GRASP}$ achieved SoTA performance in terms of the F1 and F1$_c$ scores, even though our method only leveraged a PLM without adding any extra layers. 

\section*{Acknowledgements}
This work was supported by Institute of Information \& communications Technology Planning \& Evaluation (IITP) grant funded by the Korea government (MSIT) (No. 2020-0-00368, A Neural-Symbolic Model for Knowledge Acquisition and Inference Techniques), the MSIT (Ministry of Science and ICT), Korea, under the ITRC (Information Technology Research Center) support program (IITP-2022-2018-0-01405) supervised by the IITP, and the MSIT, Korea, under the ICT Creative Consilience program (IITP-2022-2020-0-01819) supervised by the IITP.

\bibliography{anthology,custom}

\appendix\label{appendix}
\clearpage

\section{Detailed experimental settings\label{appendix:experiment_settings}}
\begin{table}[ht]
\centering
\scalebox{0.9}{
\begin{tabular}{l|cc}
\toprule
\begin{tabular}[c]{@{}l@{}}Hyper-\\ parameters\end{tabular} & \multicolumn{1}{l|}{$\mathbf{GRASP}_{base}$} & $\mathbf{GRASP}_{large}$ \\ \midrule
Learning rate   & \multicolumn{1}{c|}{$5e-5$} & $5e-6$ \\ \midrule
Max seq. len & \multicolumn{2}{c}{512} \\ \midrule
Batch size & \multicolumn{2}{c}{8} \\ \midrule
Num. epochs & \multicolumn{2}{c}{30} \\ \midrule
\begin{tabular}[c]{@{}l@{}}Joint ratio\\ ($\lambda_{1}$ \& $\lambda_{2}$)\end{tabular}    & \multicolumn{2}{c}{0.7 / 0.3} \\  \bottomrule
\end{tabular}
} \caption{Hyper-parameter values used in prompt-tuning process on test set.\label{table:hyperparams}}
\end{table}

$\mathbf{GRASP}$ is trained using AdamW~\citep{loshchilov2017decoupled} as an optimizer with no weight decay. The number of training epochs is set to 30 with early stopping, and the ratio of $\lambda_{1}$ and $\lambda_{2}$ for the joint loss is 0.7 to 0.3. A learning rate of $5e{-5}$, batch size of 8, and maximum sequence length of 512 are adopted for RoBERTa-base with identical parameters for RoBERTa-large, except for the learning rate of $5e{-6}$.
\begin{table*}[t]
\centering
\scalebox{0.8}{
\begin{tabular}{ccccc}
    \toprule
    \multicolumn{5}{l}{\textbf{Dialogue}} \\ \midrule
    \multicolumn{5}{l}{\begin{tabular}[c]{p{19cm}}
S1: Hey! Hi!\\
\textbf{S2}: Hey!\\
S1: What’s up?\\
\textbf{S2}: Well umm, \textbf{Chandler} and I are \underline{moving in together}.\\
S1: Oh my God. Ohh, my little sister and my best friend…shaking up. Oh, that’s great. That’s great.\\
S3: Guys, I’m happy too.\\
\textbf{S2}: Okay, come here!\\
S3: Wow! Big day huh? People moving in, people getting annulled…\\
\textbf{S2}: Okay, I gotta go find Rachel but umm, if you guys see her could you please try to give her some really bad news so that mine doesn’t seem so bad?\\
...
    \end{tabular}} \\ \midrule
    \multirow{2}{*}{\textbf{Argument pair}} & \multicolumn{1}{c|}{\multirow{2}{*}{\textbf{Ground Truth}}} & \textbf{RoBERTa}  & \multicolumn{2}{|c}{$\mathbf{GRASP}$} \\ & \multicolumn{1}{c|}{} & Predicted Relation & \multicolumn{1}{|c}{Predicted (\textit{subject, object, trigger})} & Predicted Relation \\ \midrule
    (\textbf{S2}, \textbf{Chandler}) & \multicolumn{1}{c|}{per:girl/boyfriend} & per:roommate & \multicolumn{1}{|c}{(S2, Chandler, \underline{moving in together})} & per:girl/boyfriend \\
    
    (\textbf{Chandler}, \textbf{S2}) & \multicolumn{1}{c|}{per:girl/boyfriend} & per:roommate & \multicolumn{1}{|c}{(Chandler, S2, \underline{moving in together})} & per:girl/boyfriend \\
    
    \bottomrule
    \end{tabular}
}
\centering
\scalebox{0.8}{
\begin{tabular}{ccccc}
    \toprule
    \multicolumn{5}{l}{\textbf{Dialogue}} \\ \midrule
    \multicolumn{5}{l}
    {\begin{tabular}[c]{p{19cm}}
S1: Rach, Rach, I just remembered. I had a dream about \textbf{Mr. Geller} last night.\\
S2: Really?!\\
S1: Yeah, I dreamt that he saved me from a burning building and he was so brave and so strong! And it’s making me look at him totally differently. Y’know, I mean he used to be just, y’know “Jack Geller \textbf{Monica} and \textbf{Ross}’s \underline{dad}” and now he’s he’s “Jack Geller, dream hunk."\\
S2: I dunno. Y’know to me he’ll always be “Jack Geller, walks in while you’re changing.”
    \end{tabular}
    } \\ \midrule
    \multirow{2}{*}{\textbf{Argument pair}} & \multicolumn{1}{c|}{\multirow{2}{*}{\textbf{Ground Truth}}} & \textbf{RoBERTa}  & \multicolumn{2}{|c}{$\mathbf{GRASP}$} \\ & \multicolumn{1}{c|}{} & Predicted Relation & \multicolumn{1}{|c}{Predicted (\textit{subject, object, trigger})} & Predicted Relation \\ \midrule
    (Mr. Geller, Monica) & \multicolumn{1}{c|}{\textit{per:children}} & unanswerable & \multicolumn{1}{|c}{(Mr Geller, Monica, \underline{dad})} & per:children \\
    (Mr. Geller, Ross) & \multicolumn{1}{c|}{\textit{per:children}} & unanswerable & \multicolumn{1}{|c}{(Mr Geller, Ross, \underline{dad})} & per:children \\
    (Ross, Mr. Geller) & \multicolumn{1}{c|}{\textit{per:parents}} & per:alternate\_names & \multicolumn{1}{|c}{(Ross, Mr Geller, \underline{dad})} & per:parents \\ \bottomrule
    \end{tabular}
    }\caption{The additional examples for qualitative analysis on the prediction of $\mathbf{GRASP}$ based on the comparison with the RoBERTa model. Predicted (\textit{subject, object, trigger}) is that $\mathbf{GRASP}$ predicted on RCD task.\label{table:appen_qual1}}
\end{table*}

\section{Qualitative Analysis Examples\label{appendix:qualitative}}
Table \ref{table:appen_qual1} shows additional examples to demonstrate the prediction tendency of $\mathbf{GRASP}$ and the fine-tuned RoBERTa models on the symmetrical relations described in Section \ref{subsec:qualitative}.

\end{document}